\documentclass[letterpaper,journal]{IEEEtran}

\usepackage{amsmath,amsfonts,amssymb,amsthm}
\usepackage{algorithm}
\usepackage{algpseudocode}
\usepackage{array}
\usepackage{booktabs}
\usepackage{multirow}
\usepackage{graphicx}
\usepackage{stfloats}
\usepackage{url}
\usepackage{cite}
\usepackage{microtype}
\usepackage{mathtools}
\usepackage{xcolor}
\usepackage{hyperref}
\usepackage[table]{xcolor}
\newcolumntype{g}{>{\columncolor{gray!12}[1.5pt][1.5pt]}c}
\begin{document}

\title{FedUDG: Federated Domain Generalization with Unlabeled Clients}

\author{Junyu Ren, Wensheng Gan, Philip S. Yu,~\IEEEmembership{Life Fellow,~IEEE}
\thanks{This research was supported in part by the National Natural Science Foundation of China (No. 62272196) and Guangzhou Basic and Applied Basic Research Foundation (No. 2024A04J9971).}
\thanks{Junyu Ren and Wensheng Gan are with the College of Cyber Security, Jinan University, Guangzhou 510632, China. (E-mail: renjunyu193@gmail.com, wsgan001@gmail.com)}
\thanks{Philip S. Yu is with the Department of Computer Science, University of Illinois Chicago, Chicago, USA. (E-mail: psyu@uic.edu)}
\thanks{Corresponding author: Wensheng Gan.}}

\markboth{IEEE Transactions on Mobile Computing, 2026}%
{Ren \MakeLowercase{\textit{et al.}}: FedUDG: Federated Domain Generalization with Unlabeled Clients}

\maketitle

\begin{abstract}
Federated domain generalization (FDG) aims to learn models that generalize reliably to out-of-distribution (OOD) environments without centralizing client's data. However, existing FDG methods assume that all source clients have labeled data, limiting their ability to leverage the abundant unlabeled client data available in practice. To address this limitation, we propose FedUDG. At the federation level, FedUDG utilizes a parameter-functional differentiation strategy that maintains a semantic anchor updated solely using ground-truth labels. The remaining parameters absorb knowledge from unlabeled domains under bounded aggregation, preventing domain-biased noise from eroding the class semantics. Within each unlabeled client, FedUDG utilizes a student-teacher framework with two complementary objectives: (1) semantic self-training on high-confidence samples, using the functional differentiation between the semantic anchor and classifier to constrain the model to learn semantically pure and cross-domain stable features; and (2) label-free representation distillation on all unlabeled samples to stabilize learning under appearance variations. Together, these objectives guide the model toward a transferable feature space, while mitigating overfitting to domain-specific characteristics. Furthermore, to address the lack of systematic evaluation of FDG under formal privacy constraints, we develop a sample-level differentially private extension, DP-FedUDG, to quantify the trade-off between privacy and OOD generalization. Across four evaluation protocols, FedUDG consistently outperforms state-of-the-art baselines. The code is available at \url{https://github.com/Jr10426/FedUDG}
\end{abstract}


\section{Introduction}
\label{sec:introduction}

Federated learning (FL) enables multiple clients to train a model collaboratively without directly exchanging raw data \cite{mcmahan2017communication}. Clients exhibit heterogeneity in data distributions, acquisition processes, and dataset sizes. Such heterogeneity can induce client drift, divergent local objectives, and slow convergence \cite{li2020fedprox,karimireddy2020scaffold}. Several optimization methods address these differences among participating clients \cite{li2020fedprox,karimireddy2020scaffold}. However, their evaluations generally focus on distributions represented during training and do not directly assess performance in unseen environments.
In deployment, changes in environments and acquisition devices can introduce distributional shifts between source and target data. Robustness to these OOD shifts is important for reliable deployment \cite{abbavaram2026boosting}. Domain generalization (DG) studies how to learn from multiple observed source domains and generalize to an unseen target domain \cite{ren2026domain,gulrajani2021domainbed}. Centralized DG usually assumes joint access to all source-domain data, which is infeasible across institutions. Federated domain generalization (FDG) instead uses variation across source clients while keeping raw data local. Existing FDG methods address domain shifts through alignment \cite{gupta2025fedalign}, data manipulation \cite{liu2021feddg}, learning strategies \cite{guo2023invariant}, and aggregation optimization \cite{zhang2023feddgga}.

Despite recent progress, FDG remains underexplored in realistic settings where annotation is scarce. Annotation typically demands domain expertise and sustained human effort, making it realistic for a few clients to provide labels while many others possess only unlabeled samples. Federated semi-supervised learning (FSSL) \cite{jeong2021fedmatch,liu2021fedirm,qiu2024fedanchor} has been proposed to address label scarcity. FSSL assumes that training and test data come from the same distribution \cite{liu2021fedirm,liang2022rscfed}, whereas FDG requires generalization to unseen out-of-distribution clients. This fundamental difference renders existing FSSL methods inapplicable to FDG tasks with unlabeled source clients: FSSL only needs to consider model performance on in-distribution clients and thus tends to learn domain-specific features during optimization. These features indeed can boost performance on unlabeled clients locally, but they do not constitute transferable class semantics and thus harm out-of-distribution generalization \cite{arjovsky2019irm}. For FDG tasks with unlabeled clients, this implies a dual requirement: we should exploit unlabeled data for learning while simultaneously constraining the influence of domain-specific information.
Furthermore, formal privacy protection remains underexplored in FDG. Several FDG studies regard local data retention as a privacy safeguard \cite{xu2024fadh,lai2025bpf}, while others report low leakage under selected inversion attacks \cite{nguyen2025pardon}. However, a rigorous privacy guarantee requires an explicit adjacency relation, a randomized mechanism, and cross-round privacy accounting \cite{abadi2016deep,mironov2017renyi,fu2026differentially}. Taken together, existing FDG research has two limitations: (1) it does not exploit unlabeled clients to improve OOD generalization; and (2) it does not evaluate generalization under formal privacy constraints.

\begin{figure}[t]
    \centering
    \includegraphics[width=0.6\columnwidth]{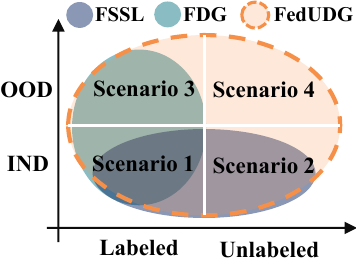}
    \caption{Comparison of the problem settings of FSSL, FDG, and this work. FSSL jointly exploits labeled and unlabeled data but typically evaluates on the client distributions seen during training. FDG targets OOD generalization to unseen target clients but generally assumes that all source clients are labeled. FedUDG addresses both heterogeneous label availability across source clients and generalization to an unseen OOD target client.}
    \label{fig:scope}
\end{figure}

To address these challenges, we propose FedUDG, a novel approach for FDG with unlabeled clients that bridges the gap illustrated in Fig.~\ref{fig:scope}. Our method operates through two core mechanisms: (1) Federated-level semantic protection. We introduce a parameter-functional differentiation strategy. Specifically, we maintain a semantic anchor updated exclusively using ground-truth labels, serving as a stable reference for class semantics. The remaining parameters, including the classifier, can incorporate knowledge from unlabeled clients. The core rationale for this design is as follows: although unlabeled data carries rich cross-domain knowledge, noise in supervisory signals introduced by domain bias can compromise the model. By partitioning parameters into two categories: those maintained solely by ground-truth labels and those receiving knowledge from unlabeled clients, we enable the global update to leverage information from unlabeled clients while remaining constrained within the boundaries by genuine class semantics. 
(2) Semi-supervised representation learning with transferability constraints. To provide stable and consistent learning objectives for unlabeled clients, we adopt a student-teacher framework. Building upon the parameter-functional differentiation strategy, we supervise the student model via a consensus mechanism derived from the teacher: a sample receives supervisory signals to guide student learning only when the semantic anchor and classifier agree on the same class with high confidence and maintain consistent predictions under independent perturbations. The core insight is that the two predictors maintain fundamentally different supervisory characteristics: the semantic anchor represents a pure semantic baseline, while the classifier carries cross-domain bias. Their consensus helps ensure that the supervisory targets are both semantically faithful and stable across domain variations. This supervision strategy guides the model's learning process toward a transferable feature space, mitigating overfitting to domain-specific cues. Concurrently, for all unlabeled samples, we employ label-free representation distillation, aligning the student's strongly augmented representations with the teacher's weakly augmented ones. This objective encourages consistency and robustness in the representation space, enabling the model to make use of all unlabeled data while reducing sensitivity to domain-related appearance variations.

Based on this framework, we extend FedUDG with sample-level differential privacy, resulting in DP-FedUDG, which allows us to quantify the trade-off between privacy and OOD generalization. Our contributions are summarized as follows.

\begin{itemize}
    \item To the best of our knowledge, this is the first study of federated domain generalization with unlabeled clients.
    \item We propose a parameter-functional differentiation strategy that maintains a semantic anchor updated solely using ground-truth labels as a stable reference, enabling the model to absorb cross-domain knowledge from unlabeled clients while remaining aligned with genuine class semantics. This design allows the model to effectively leverage unlabeled clients for FDG.
    \item We introduce a student-teacher framework with two objectives: semantic self-training on credible samples that leverages the distinct roles of the semantic anchor and classifier to learn semantically faithful and cross-domain stable features, and representation distillation on all unlabeled samples to stabilize learning under appearance variations without semantic supervision. These objectives guide the model toward a transferable feature space, mitigating overfitting to domain-specific characteristics. 
    \item We develop DP-FedUDG, a sample-level differentially private extension, and evaluate the trade-off between privacy budget and OOD generalization under an explicit adjacency relation and cross-round privacy accounting.
    \item Across four evaluation protocols, FedUDG achieves the highest mean accuracy on unseen target clients in 12 of the 13 evaluated settings.
\end{itemize}

The remainder of this paper is organized as follows. Section~\ref{sec:related} reviews related work. Section~\ref{sec:method} presents FedUDG. Section~\ref{sec:experiments} reports the experimental results and analysis. Section~\ref{sec:conclusion} concludes the paper and discusses future research directions.

\section{Related Work}
\label{sec:related}

This section reviews related work on FedUDG, including federated domain generalization, federated semi-supervised learning, and privacy protection in FDG.

\subsection{Federated Domain Generalization}

Federated domain generalization \cite{li2025federated} trains a model collaboratively on multiple source clients without exchanging raw data and deploys it to a target client that is unseen during training. Unlike conventional FL, which focuses on average performance over participating distributions, FDG evaluates generalization to a non-participating target distribution. Existing FDG methods fall into four main categories \cite{li2025federated}:
(1) \emph{Federated domain alignment methods} reduce discrepancies among client representations or prediction distributions without sharing raw samples. Representative implementations include statistical distance constraints \cite{sun2023feature}, adversarial alignment \cite{zhang2023iot}, and cross-client feature alignment \cite{gupta2025fedalign}. FedAlign \cite{gupta2025fedalign} transfers and aligns features across clients so that each local model observes representation variations from other source domains. These methods use a shared representation space to exchange knowledge and promote more consistent decision structures across clients.
(2) \emph{Data-manipulation methods} expand the observed support of source domains to simulate potential distribution shifts. Common strategies include data generation \cite{zhang2025causality}, data augmentation \cite{liu2021feddg,chen2023ccst}, and normalization \cite{yuan2023collaborative}. FedDG \cite{liu2021feddg} exchanges and interpolates amplitude information in the frequency domain and uses episodic learning to simulate continuous domain shifts. CCST \cite{chen2023ccst} exchanges client-level channel-style statistics and transfers them to labeled samples for cross-client style augmentation. 
(3) \emph{Learning-strategy methods} use local objectives and training procedures to learn transferable representations. They include domain-invariant representation learning \cite{liao2025decision,guo2023invariant}, adversarial learning \cite{zhang2023iot}, feature disentanglement \cite{zhang2025causality}, meta-learning \cite{liu2021feddg}, knowledge distillation \cite{liu2026knowledge}, and invariant-risk regularization \cite{guo2023invariant,arjovsky2019irm}. Representation learning and feature disentanglement respectively model shared discriminative structure and domain-specific factors. Meta-learning and episodic training construct simulated domain-transfer tasks across source clients. 
(4) \emph{Aggregation-optimization methods} adjust how local models, gradients, or client weights are combined at the server \cite{zhang2023feddgga,nguyen2025fedomg}. FedDG-GA \cite{zhang2023feddgga} adjusts client contributions according to generalization gaps. FedOMG \cite{nguyen2025fedomg} matches client gradients and balances the average descent direction with cross-domain consistency.

These methods generally assume that all source clients are labeled and rely on ground-truth supervision to learn transferable features. They do not consider source clients without labels, where unlabeled supervision may introduce pseudo-label noise and domain-specific shortcuts. FedUDG extends this setting to labeled and unlabeled source clients while retaining evaluation on an unseen OOD target.

\subsection{Federated Semi-Supervised Learning}

Federated semi-supervised learning \cite{jeong2021fedmatch} studies collaborative learning when raw data are decentralized and labels are incomplete. Where labels reside determines both the supervision available to the algorithm and the paths through which pseudo-label errors propagate. Existing work can be organized into three settings according to client-level label topology:
(1) Under \emph{intra-client partial labeling},  every client contains a small labeled subset and a larger unlabeled subset. FedMatch \cite{jeong2021fedmatch} uses parameter decomposition and inter-client consistency to reduce interference between supervised and unsupervised objectives. SAGE \cite{liu2025sage}  exploits confidence discrepancies between local and global models to correct mismatched pseudo-labels, and ProxyFL \cite{chen2026proxyfl} uses classifier proxies and a cross-client proxy pool to recover low-confidence samples. Since each client has some ground-truth labels, local supervision can directly calibrate pseudo-label thresholds, class priors, and decision boundaries.
(2) Under \emph{inter-client label separation}, some clients are fully labeled while the others are fully unlabeled, so unlabeled clients receive no local supervision. FedIRM \cite{liu2021fedirm} extracts class relations from labeled clients and matches them on unlabeled clients. RSCFed \cite{liang2022rscfed} samples client models to construct multiple sub-consensus models and reweights them by model distance. XCLP \cite{scott2023xclp} performs label propagation over a cross-client data graph. FedRVR \cite{yang2025fedrvr} models data and model relations and uses relation-guided aggregation to reduce the influence of biased unlabeled clients, and FedPDM \cite{liao2025fedpdm} progressively matches pseudo-label distributions through optimal transport. These methods mainly stabilize the use of fully unlabeled clients through relational, consensus-based, or distributional constraints.
(3) Under \emph{server-side labeling}, a small labeled set resides on the server and all clients are unlabeled. SemiFL \cite{diao2022semifl} alternates between supervised server fine-tuning and client-side pseudo-label training. $(FL)^2$ \cite{lee2024fl2} combines adaptive thresholds, sharpness-aware consistency, and learning-state-aware aggregation, and BSemiFL \cite{wang2025bsemifl} fuses global and local predictions in a Bayesian manner to generate pseudo-labels. In this setting, the server-labeled set provides a common reference for checking and correcting client-generated pseudo-labels.

Inter-client label separation is closest to the data topology studied here, but existing methods in this setting mainly optimize semi-supervised classification over the participating client distributions and do not evaluate generalization to an unseen target client \cite{liu2021fedirm,liang2022rscfed,yang2025fedrvr}. Under source--target shifts, even correctly predicted pseudo-labels may reinforce domain-specific shortcuts, and unlabeled updates may propagate the resulting biases into the global model. FedUDG adopts the same label topology but targets generalization capability: it learns from unlabeled clients through semantic self-training and representation distillation that encourage cross-domain semantic consistency and stability under appearance perturbations.

\subsection{Privacy in Federated Domain Generalization}

Several FDG studies frame privacy in terms of local data retention and restricted information exchange. FADH \cite{xu2024fadh} keeps its domain hallucinator local and exchanges nothing beyond model weights, and BPF \cite{lai2025bpf} communicates through client-side proxy models and transfers knowledge by mutual distillation. These designs avoid exposing raw samples or explicit domain statistics, but the server still observes model or proxy-model updates, from which training records may be reconstructed \cite{zhu2019deep}.
Other methods replace sample exchange with compact statistics or parameters. CCST \cite{chen2023ccst} transmits channel-wise style statistics, FedAlign \cite{gupta2025fedalign} shares feature means and variances for cross-client feature expansion, and PLAN \cite{gong2026plan} encodes domain knowledge in lightweight visual and textual prompts that are exchanged and aggregated. PARDON \cite{nguyen2025pardon} uploads client-level style statistics to construct interpolated global styles, and empirically evaluates style leakage through generative reconstruction attacks. Such abstraction reduces direct information exposure and can resist selected inversion attacks \cite{nguyen2025pardon}.
However, these methods assess privacy through data localization, information abstraction, or attack experiments. They typically neither specify a sample-level adjacency relation nor account for privacy loss across communication rounds, and therefore do not provide a composable $(\varepsilon,\delta)$ guarantee. To examine unseen-domain generalization under formal privacy constraints, we develop DP-FedUDG, which applies sample-level differential privacy with cross-round privacy accounting to FedUDG.

\section{Proposed Method}
\label{sec:method}

This section defines the task and presents FedUDG's learning objectives, model aggregation, and privacy extension.

\begin{figure*}[t]
    \centering    \includegraphics[width=0.95\textwidth]{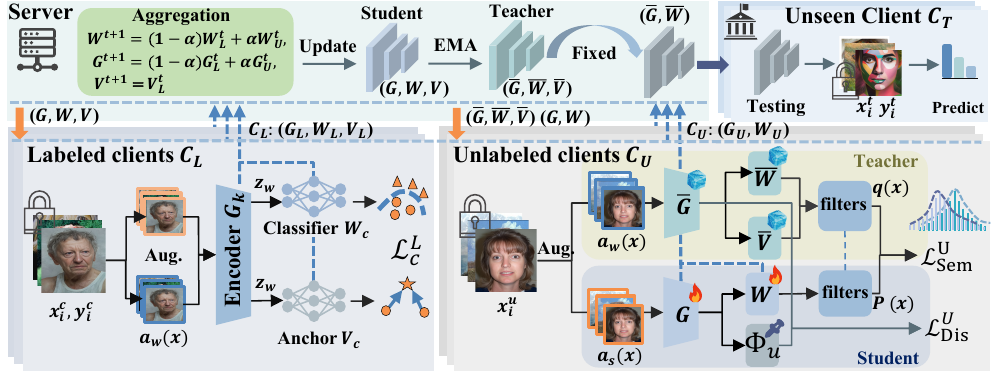}
    \caption{Overall architecture of FedUDG. Labeled clients learn the encoder, classifier, and semantic anchors from ground-truth labels. Within a student-teacher framework, unlabeled clients perform semantic self-training on credible samples and label-free representation distillation on all samples. The server follows the parameter-functional differentiation strategy when aggregating model components and updates the EMA teacher.}
    \label{fig:framework}
\end{figure*}

\subsection{Problem Formulation and Method Overview}

Suppose that $M$ source clients participate in federated training. Their set is $\mathcal C_S=\mathcal C_L\cup\mathcal C_U$, where $\mathcal C_L$ contains $M_L=|\mathcal C_L|$ fully labeled clients and $\mathcal C_U$ contains $M_U=|\mathcal C_U|$ fully unlabeled clients. We assume $M=M_L+M_U$, $1\leq M_L<M_U$, and $\mathcal C_L\cap\mathcal C_U=\varnothing$. Each source client $k\in\mathcal C_S$ represents a training environment with joint distribution $P_k(X,Y)$. All clients share the input space $\mathcal X$ and label space $\mathcal Y=\{1,\ldots,K\}$, but their data distributions may differ. A labeled client $c\in\mathcal C_L$ holds $\mathcal S_c^L=\{(x_i^c,y_i^c)\}_{i=1}^{n_c}$ sampled independently from $P_c(X,Y)$. An unlabeled client $u\in\mathcal C_U$ observes only $\mathcal S_u^U=\{x_i^u\}_{i=1}^{n_u}$ sampled from the marginal distribution $P_u(X)$, and its class labels are unavailable during training.
The unseen target client is denoted by $C_T$, where $C_T\notin\mathcal C_S$. Its joint distribution differs from those of all source clients: $P_T(X,Y)\neq P_k(X,Y)$ for every $k\in\mathcal C_S$. The target test set $\mathcal S_T=\{(x_i^t,y_i^t)\}_{i=1}^{n_t}$ is used only after training. Given $\{\mathcal S_c^L\}_{c\in\mathcal C_L}$ and $\{\mathcal S_u^U\}_{u\in\mathcal C_U}$, the goal is to learn a predictor $f:\mathcal X\rightarrow\Delta^{K-1}$ through federated collaboration without exchanging raw data. Its target risk is $R_T(f)=\mathbb E_{(x,y)\sim P_T}[\ell(f(x),y)]$, where $\ell$ is the per-sample classification loss.

Fig.~\ref{fig:framework} shows the FedUDG architecture. FedUDG combines a parameter-functional differentiation strategy with a student-teacher framework for learning from unlabeled clients. Labeled data provide class-semantic supervision, whereas unlabeled data provide cross-domain representation signals. The server maintains a student model and an exponential moving average (EMA) teacher \cite{tarvainen2017mean}. The student consists of an encoder $G:\mathcal X\rightarrow\mathbb R^d$, a linear classifier $W\in\mathbb R^{K\times d}$, and cosine semantic anchors $V\in\mathbb R^{K\times d}$. The corresponding teacher parameters are $(\bar G,\bar W,\bar V)$. FedUDG first uses ground-truth labels to establish the shared representation, linear decision boundary, and semantic anchors. The EMA teacher provides consensus-based soft semantic targets for semantic self-training on credible samples and representation targets for label-free representation distillation on all unlabeled samples. Each unlabeled client $u$ retains a local projection head $\Phi_u$. The server coordinates the two learning paths and updates the temporally smoothed teacher. Unlabeled clients contribute updates to the shared encoder and classifier, while labeled clients update the semantic anchors. After training, the EMA checkpoint with the best labeled-source validation score is selected, and inference uses only $\bar G$ and $\bar W$.

\subsection{Supervised Learning on Labeled Clients}

Labeled clients form the supervised semantic-learning path and train the encoder, linear classifier $W$, and semantic anchors with ground-truth labels. For an encoder feature $z=G(x)$, the linear classifier produces
$p_W(\cdot\mid x;G,W)=\operatorname{softmax}(Wz)$. The semantic-anchor matrix is $V=[v_1^\top;\ldots;v_K^\top]$. Let $\epsilon_0>0$ be a numerical-stability constant, $\hat z=z/(\lVert z\rVert_2+\epsilon_0)$, and $\hat v_j=v_j/(\lVert v_j\rVert_2+\epsilon_0)$. Anchor predictions are given by
$p_V(y=j\mid x;G,V)=\exp(\hat v_j^\top\hat z/\beta_V)/\sum_{r=1}^{K}\exp(\hat v_r^\top\hat z/\beta_V)$, where $\beta_V>0$ is the anchor temperature. The linear classifier learns a decision boundary from unnormalized features, whereas anchor predictions provide a geometric semantic reference through the cosine similarity between feature directions and class directions.

For a labeled sample $(x,y)$, the client draws one view $a_w^1(x)$ from the weak augmentation distribution $\mathcal T_w$ and obtains $z_w=G(a_w^1(x))$; the superscript indexes an independent augmentation draw. The same feature is passed to $W$ and $V$. A labeled client $c\in\mathcal C_L$ minimizes
\begin{equation}
\begin{aligned}
\mathcal L_c^L
={}&\mathbb E_{(x,y)\sim\mathcal S_c^L}\big[
\operatorname{CE}\!\left(p_W(\cdot\mid a_w^1(x)),y\right)\\
&+\lambda_V\operatorname{CE}\!\left(p_V(\cdot\mid a_w^1(x)),y\right)
\big],
\end{aligned}
\label{eq:labeled}
\end{equation}
where $\lambda_V$ weights the anchor loss. At round $t$, every labeled client starts from the same global student state and obtains $(G_c^{(t,+)},W_c^{(t,+)},V_c^{(t,+)})$ after local optimization, with each transmitted anchor row normalized to unit norm. The superscript $(t,+)$ denotes a state after local training in round $t$ and before the next global update. The server averages corresponding parameters equally across labeled clients, giving $(G_L^{(t,+)},W_L^{(t,+)},V_L^{(t,+)})$. Unlabeled training begins after supervised warm-up to reduce reliance on poorly calibrated early teacher predictions. During joint training, labeled clients continue to optimize Eq. \eqref{eq:labeled} in each round before their updates are fused with those from unlabeled clients.

\subsection{Unlabeled-Client Learning}

Each unlabeled client receives the current student and EMA teacher. The teacher produces semantic evidence and representation targets from weakly augmented inputs, while the student learns from strongly augmented inputs. This student-teacher framework combines two objectives: semantic self-training on credible samples selected by teacher consensus and label-free representation distillation on all unlabeled samples.

\subsubsection{Teacher consensus and soft-target construction}

For an unlabeled sample $x$, the client draws two independent weak augmentations, $a_w^1(x)$ and $a_w^2(x)$, and one strong augmentation $a_s(x)$. The weakly augmented inputs are passed through the teacher encoder and classifier to obtain $p_{W,1}=p_W(\cdot\mid a_w^1(x);\bar G,\bar W)$ and $p_{W,2}=p_W(\cdot\mid a_w^2(x);\bar G,\bar W)$. The teacher feature of the first weakly augmented input is also passed to the teacher anchors, giving $p_{V,1}=p_V(\cdot\mid a_w^1(x);\bar G,\bar V)$.
For each prediction $p_b$, let $\hat y_b=\arg\max_j p_{b,j}$ and $c_b=\max_j p_{b,j}$. A sample is admitted only when three conditions hold. Class agreement requires $\hat y_{W,1}=\hat y_{W,2}=\hat y_{V,1}$ and excludes explicit semantic conflicts. This condition cross-checks the linear classifier against cosine semantic anchors maintained by labeled-client supervision. The confidence condition requires $\min\{c_{W,1},c_{W,2},c_{V,1}\}\geq\tau$, preventing low-confidence predictions from passing solely because their predicted classes agree. For distributional consistency, let $\mathcal P=(p_{W,1},p_{W,2},p_{V,1})$ be the ordered collection of three predictions and $\tilde p=|\mathcal P|^{-1}\sum_{p\in\mathcal P}p$. The sample must further satisfy $\operatorname{JS}(\mathcal P)=|\mathcal P|^{-1}\sum_{p\in\mathcal P}\operatorname{KL}(p\|\tilde p)\leq\kappa$, which limits disagreement among the full predictive distributions beyond the top-1 class. Because the linear classifier and anchors receive different supervision, their consensus selects samples with confident, mutually consistent semantic predictions for self-training.
Let $m(x)\in\{0,1\}$ indicate whether all three conditions hold. For an admitted sample, the classwise log-consensus score is $s_j(x)=|\mathcal P|^{-1}\sum_{p\in\mathcal P}\log(p_j+\epsilon_0)$. The normalized geometric mean then defines the soft semantic target
\begin{equation}
q_j(x)=\frac{\exp\{s_j(x)\}}
{\sum_{r=1}^{K}\exp\{s_r(x)\}}.
\label{eq:softtarget}
\end{equation}
The geometric mean reduces the weight of any class that receives a low probability from one prediction. It is computed in log space to avoid numerical underflow. The full probability vector $q(x)$ integrates the three teacher predictions and provides semantic supervision for the strongly augmented input.

\subsubsection{Semantic self-training for admitted samples}

Admitted samples provide semantic supervision through the EMA teacher's consensus soft semantic target. The student matches this target under strong augmentation, encouraging class-consistent predictions across source domains and prediction stability under the applied appearance perturbations. The student prediction on $a_s(x)$ is $p_W(\cdot\mid a_s(x);G,W)$. The teacher-derived $m(x)$ and $q(x)$ are treated as constants during local optimization. For a mini-batch $\mathcal B_u$ on unlabeled client $u$, the semantic loss is
\begin{equation}
\mathcal L_{\mathrm{sem}}^U=
\frac{
\sum_{x\in\mathcal B_u}m(x)
\operatorname{SCE}\!\left(p_W(\cdot\mid a_s(x);G,W),q(x)\right)}
{\max\!\left(1,\sum_{x\in\mathcal B_u}m(x)\right)},
\label{eq:semantic}
\end{equation}
where $\operatorname{SCE}(p,q)=-\sum_{j=1}^{K}q_j\log p_j$ is soft-target cross-entropy. Normalizing by the number of admitted samples reduces changes in loss scale across mini-batches with different admission rates. If no sample is admitted, $\mathcal L_{\mathrm{sem}}^U$ is set to zero. This design follows the weak-to-strong augmentation consistency principle of FixMatch \cite{sohn2020fixmatch}, while combining predictions from the classifier and semantic anchors.

\subsubsection{Label-free representation distillation}

To exploit all unlabeled samples, FedUDG complements semantic self-training with label-free representation distillation. The teacher provides representation targets for every sample, including those not admitted for semantic self-training. For the same sample, the EMA teacher extracts a stable target representation from a weakly augmented input, while the student encoder extracts a representation from a strongly augmented input. A local projection head $\Phi_u:\mathbb R^d\rightarrow\mathbb R^d$ maps the student representation into the teacher feature space. The distillation loss is
\begin{equation}
\mathcal L_{\mathrm{Dis}}^U=
\frac{1}{|\mathcal B_u|}
\sum_{x\in\mathcal B_u}
\left[1-
\cos\!\left(
\Phi_u(G(a_s(x))),
\operatorname{sg}[\bar G(a_w^1(x))]
\right)\right],
\label{eq:Dis}
\end{equation}
where $\cos(\cdot,\cdot)$ denotes cosine similarity and $\operatorname{sg}$ stops gradients. The fixed target branch provides a stable reference during local optimization, while the local projection head maps the student's representation of the strongly augmented input toward the teacher's representation of the weakly augmented input. This asymmetric design is inspired by SimSiam \cite{chen2021simsiam} and uses an EMA teacher to provide temporally smoothed targets. Equation~\eqref{eq:Dis} enforces consistency in representation space using teacher features as targets, independently of class-semantic supervision. This consistency stabilizes feature learning under the applied appearance perturbations and promotes transfer to unseen domains.

\subsubsection{Local unlabeled objective}

The two unlabeled-client losses are combined with a proximal regularizer to control client drift:
\begin{equation}
\mathcal L_u^U=
\lambda_{\mathrm{Dis}}\mathcal L_{\mathrm{Dis}}^U
+\lambda_{\mathrm{sem}}(t)\mathcal L_{\mathrm{sem}}^U
+\frac{\mu}{2}
\left\lVert
\operatorname{par}(G_u)-\operatorname{par}(G^{(t)})
\right\rVert_2^2,
\label{eq:unlabeled}
\end{equation}
where $\lambda_{\mathrm{Dis}}$ controls representation distillation, $\lambda_{\mathrm{sem}}(t)$ controls semantic self-training, and $\mu\geq0$ is the proximal coefficient. Here, $\operatorname{par}(G)$ is the vector of trainable encoder parameters. The proximal term uses the global encoder $G^{(t)}$ distributed at the beginning of round $t$ as its reference. It limits excessive drift of $G_u$ on heterogeneous unlabeled data \cite{li2020fedprox}.

Let $R_{\mathrm{warm}}$ be the number of warm-up rounds and $R_{\mathrm{ramp}}$ be the ramp-up duration. The semantic weight is $\lambda_{\mathrm{sem}}(t)=\lambda_{\mathrm{sem}}^{\max}\eta(t)$, where
\begin{equation}
\eta(t)=
\begin{cases}
0, & t<R_{\mathrm{warm}},\\
\min\!\left(1,\dfrac{t-R_{\mathrm{warm}}+1}{R_{\mathrm{ramp}}}\right),
& t\geq R_{\mathrm{warm}},
\end{cases}
\label{eq:ramp}
\end{equation}
Representation distillation remains active throughout joint training, while the ramp-up limits the influence of early teacher predictions. Together, the two objectives encourage cross-domain semantic consistency and stabilize discriminative learning under appearance perturbations, supporting generalization to unseen domains.

\subsection{Model Aggregation and EMA Teacher Update}

Following the parameter-functional differentiation strategy, the server aggregates model components from the two client groups separately after local training. Labeled clients upload $G$, $W$, and $V$. Unlabeled clients upload $G$ and $W$ and retain $\Phi_u$ locally across rounds. Therefore, the encoder and classifier incorporate updates from both groups, whereas the anchors are supervised only by ground-truth labels from labeled clients. Let $(G_L^{(t,+)},W_L^{(t,+)})$ and $(G_U^{(t,+)},W_U^{(t,+)})$ denote the group states after equal-client averaging at round $t$. The labeled-group anchor is $V_L^{(t,+)}=\operatorname{RowNorm}\!\left(M_L^{-1}\sum_{c\in\mathcal C_L}V_c^{(t,+)}\right)$, where $\operatorname{RowNorm}(\cdot)$ normalizes each row to unit $\ell_2$ norm. The server fuses the encoder and classifier with a bounded coefficient and updates the student anchors only from the labeled group:
\begin{equation}
\begin{aligned}
G^{(t+1)}&=(1-\alpha_U(t))G_L^{(t,+)}+\alpha_U(t)G_U^{(t,+)},\\
W^{(t+1)}&=(1-\alpha_U(t))W_L^{(t,+)}+\alpha_U(t)W_U^{(t,+)},\\
V^{(t+1)}&=V_L^{(t,+)}.
\end{aligned}
\label{eq:aggregation}
\end{equation}
Here, $\alpha_U(t)=\alpha_U^{\max}\eta(t)$ and $0\leq\alpha_U^{\max}<1$. Averaging within each group prevents the more numerous unlabeled clients from receiving greater weight solely because of their number. The bound on $\alpha_U(t)$ controls the weight of potentially biased unlabeled updates in the global encoder and classifier.

After obtaining the global student for round $t+1$, the server updates the teacher by EMA. The teacher anchor $\bar V^{(t+1)}$ is normalized separately after mixing the previous teacher anchor with the newly aggregated student anchor:
\begin{equation}
\begin{aligned}
\bar G^{(t+1)}&=\rho_t\bar G^{(t)}+(1-\rho_t)G^{(t+1)},\\
\bar W^{(t+1)}&=\rho_t\bar W^{(t)}+(1-\rho_t)W^{(t+1)},\\
\bar V^{(t+1)}&=\operatorname{RowNorm}\!\left[
\rho_t\bar V^{(t)}+(1-\rho_t)V^{(t+1)}
\right],
\end{aligned}
\label{eq:ema}
\end{equation}
where $\rho_t\in[0,1)$ is the EMA decay at round $t$. For a base per-step decay $\rho_0$, we use $\rho_t=\rho_0^{s_t}$, where $s_t$ is the rounded mean number of labeled-client local optimization steps in round $t$, lower-bounded by one. The server updates the teacher once after each global aggregation. Temporal averaging provides more stable semantic and representation targets than the instantaneous student \cite{tarvainen2017mean}. Teacher parameters remain fixed during local training, and $\Phi_u$ is neither aggregated nor included in the EMA update.

\subsection{Sample-Level Differential Privacy Extension}

Generalization under formal privacy constraints remains underexplored in FDG. We therefore extend FedUDG with sample-level differential privacy \cite{fu2026differentially} and call the resulting method DP-FedUDG.

\subsubsection{Threat model and privacy unit}

DP-FedUDG considers an honest-but-curious server that follows the training protocol but may inspect client messages. Clients are assumed to execute local optimization as specified. The privacy unit is one local training sample. Let $\mathcal S\simeq\mathcal S'$ denote two adjacent datasets on the same client that differ by the addition or removal of one sample. A randomized client-training mechanism $\mathsf{Priv}$ satisfies sample-level $(\varepsilon,\delta)$-differential privacy if, for every measurable output event $\mathcal O$,
\begin{equation}
\Pr[\mathsf{Priv}(\mathcal S)\in\mathcal O]
\leq e^\varepsilon
\Pr[\mathsf{Priv}(\mathcal S')\in\mathcal O]+\delta.
\label{eq:dpdef}
\end{equation}
The probability is over the internal randomness of the mechanism. $\varepsilon$ bounds the distinguishability of output distributions induced by adjacent datasets, and $\delta$ is the permitted failure probability. This definition bounds the influence of any one training sample on both the client update and the final model derived from such updates \cite{fu2026differentially}.

\subsubsection{DP-compatible masked objective}

The non-private semantic loss is normalized by the number of admitted samples. This data-dependent denominator makes each sample's gradient scale depend on other admissions in the mini-batch. DP-FedUDG constructs the private data objective from separable semantic and distillation terms. Let $p_i=p_W(\cdot\mid a_s(x_i);G,W)$ and $q_i=q(x_i)$, and let $\ell_{\mathrm{Dis},i}$ denote the cosine-distance summand in Eq.~\eqref{eq:Dis}. The unreduced loss for sample $x_i$ on an unlabeled client is
\begin{equation}
\ell_i^U=
\lambda_{\mathrm{Dis}}\ell_{\mathrm{Dis},i}
+\lambda_{\mathrm{sem}}(t)m(x_i)\operatorname{SCE}(p_i,q_i).
\label{eq:dpsemantic}
\end{equation}
The admission mask, soft semantic target, and teacher representation are computed using $x_i$ and the fixed EMA teacher. Each sample therefore contributes independently, and rejected samples retain their distillation term. Both loss terms share the expected-batch normalization applied after per-sample clipping and noise addition in Eq.~\eqref{eq:dpsgd}. On a labeled client, the unreduced per-sample loss is the summand of Eq.~\eqref{eq:labeled}, $\ell_i^L=\operatorname{CE}(p_W(\cdot\mid a_w^1(x_i)),y_i)+\lambda_V\operatorname{CE}(p_V(\cdot\mid a_w^1(x_i)),y_i)$. The encoder proximal term is applied separately as a deterministic update after the private data-gradient step.

\subsubsection{Client-side DP-SGD and privacy accounting}

Clients use Poisson sampling, independently selecting each local record at every step with sampling rate $q_k$. This sampling mechanism provides privacy amplification and supports explicit privacy accounting \cite{abadi2016deep}. Let $k\in\mathcal C_S$ be any source client and let $\operatorname{par}_k$ denote its optimizable parameter vector. For $c\in\mathcal C_L$, this vector concatenates the parameters of $(G_c,W_c,V_c)$. For $u\in\mathcal C_U$, it concatenates those of $(G_u,W_u,\Phi_u)$. For unreduced per-sample loss $\ell_i\in\{\ell_i^L,\ell_i^U\}$, DP-SGD \cite{abadi2016deep} computes $\boldsymbol\xi_i=\nabla_{\operatorname{par}_k}\ell_i$, clips it by the global $\ell_2$ norm, and adds isotropic Gaussian noise to the sum of clipped gradients:
\begin{equation}
\tilde{\boldsymbol\xi}=
\frac{1}{B_k^{\mathrm{exp}}}\left[
\sum_{x_i\in\mathcal B_k}
\boldsymbol\xi_i
\min\!\left(1,\frac{C}{\lVert\boldsymbol\xi_i\rVert_2}\right)
+\mathcal N(0,\sigma^2C^2I)
\right].
\label{eq:dpsgd}
\end{equation}
Here, $B_k^{\mathrm{exp}}=q_k n_k$ is the expected Poisson batch size used by the DP optimizer, $C$ is the clipping threshold, $\sigma$ is the noise multiplier, and $I$ is the identity matrix with the same dimension as the parameter vector. The realized size $|\mathcal B_k|$ is random and does not replace $B_k^{\mathrm{exp}}$ in the noised-gradient normalization. Clipping bounds the contribution of one sample to an update, while $\sigma$ controls the privacy--utility trade-off. BatchNorm in training mode couples samples through batch statistics. DP training therefore freezes both the running statistics and affine parameters of the pretrained BatchNorm layers. Although the local projection head is not communicated, its state affects shared updates in later rounds. It is consequently included with the encoder and classifier in per-sample clipping and noise addition.
Each client maintains a separate R\'{e}nyi differential privacy (RDP) accountant \cite{mironov2017renyi}. The accountant accumulates R\'{e}nyi privacy loss using the actual Poisson sampling rate, the number of local optimization steps, and the number of participating rounds, and converts it to $(\varepsilon_k,\delta)$-DP for a specified $\delta$ \cite{fu2026differentially}. Privacy losses across local steps and communication rounds on the same client compose sequentially. Because clients hold disjoint records, parallel composition applies across clients, and the system-level sample privacy budget is $\varepsilon_{\max}=\max_{k\in\mathcal C_S}\varepsilon_k$. Server aggregation, EMA updates, and inference are post-processing operations on privatized updates and consume no additional privacy budget.

\subsection{End-to-End Training and Inference}

\begin{algorithm}[t]
\caption{Training of FedUDG}
\label{alg:fedudg}
\footnotesize
\begin{algorithmic}[1]
\Statex \textbf{Input:} Client datasets $\{\mathcal S_c^L\}_{c\in\mathcal C_L}$ and $\{\mathcal S_u^U\}_{u\in\mathcal C_U}$;
\Statex \hspace{\algorithmicindent}warm-up rounds $R_{\mathrm{warm}}$; total rounds $R_{\mathrm{total}}$.
\Statex \textbf{Output:} Selected EMA encoder $\bar G_{\mathrm{best}}$ and classifier $\bar W_{\mathrm{best}}$.
\State Initialize the global student $(G^{(0)},W^{(0)},V^{(0)})$;
\State Set $(\bar G^{(0)},\bar W^{(0)},\bar V^{(0)})\gets(G^{(0)},W^{(0)},V^{(0)})$;
\State Initialize local projection heads $\{\Phi_u\}_{u\in\mathcal C_U}$.
\For{$t=0,\ldots,R_{\mathrm{total}}-1$}
    \State \textbf{Server:} broadcast the student to labeled clients;
    \ForAll{labeled clients $c\in\mathcal C_L$ in parallel}
        \State Set $(G_c,W_c,V_c)\gets(G^{(t)},W^{(t)},V^{(t)})$;
        \State Draw one weak augmentation per sample;
        \State Optimize $(G_c,W_c,V_c)$ using Eq.~\eqref{eq:labeled};
        \State Upload $(G_c,W_c,V_c)$ to the server.
    \EndFor
    \State \textbf{Server:} average models from $\mathcal C_L$ into $(G_L,W_L,V_L)$; normalize $V_L$.
    \If{$t\geq R_{\mathrm{warm}}$}
        \State Set $\lambda_{\mathrm{sem}}(t)$ and $\alpha_U(t)$ using Eq.~\eqref{eq:ramp};
        \State \textbf{Server:} send student and teacher to unlabeled clients;
        \ForAll{unlabeled clients $u\in\mathcal C_U$ in parallel}
            \State Set $(G_u,W_u)\gets(G^{(t)},W^{(t)})$;
            \State Draw $a_w^1(x)$, $a_w^2(x)$, and $a_s(x)$ for each sample;
            \State Compute teacher predictions $p_{W,1},p_{W,2},p_{V,1}$;
            \State Obtain $m(x)$ by agreement, confidence, and JS tests;
            \State Compute $q(x)$ for admitted samples by Eq.~\eqref{eq:softtarget};
            \State Compute $\mathcal L_{\mathrm{sem}}^U$ and $\mathcal L_{\mathrm{Dis}}^U$ by Eqs.~\eqref{eq:semantic} and \eqref{eq:Dis};
            \State Optimize $(G_u,W_u,\Phi_u)$ using Eq.~\eqref{eq:unlabeled};
            \State Upload $(G_u,W_u)$; retain $\Phi_u$ locally.
        \EndFor
        \State \textbf{Server:} equally average unlabeled models; apply Eq.~\eqref{eq:aggregation}.
    \Else
        \State \textbf{Server:} use labeled-group averages as the next student.
    \EndIf
    \State \textbf{Server:} update the EMA teacher using Eq.~\eqref{eq:ema};
    \State Evaluate the EMA model on labeled-source validation sets;
    \State Retain $(\bar G_{\mathrm{best}},\bar W_{\mathrm{best}})$ if the mean score improves.
\EndFor
\State \Return $(\bar G_{\mathrm{best}},\bar W_{\mathrm{best}})$.
\end{algorithmic}
\end{algorithm}

As summarized in Algorithm~\ref{alg:fedudg}, FedUDG begins with a supervised warm-up phase using only labeled clients, followed by joint training with labeled and unlabeled clients. During joint training, labeled clients continue supervised optimization, while unlabeled clients optimize the semantic self-training and label-free representation distillation objectives. At the start of each joint-training round, every participating client initializes its shared trainable parameters from the current global student. On unlabeled clients, the EMA teacher provides semantic and representation targets and remains fixed throughout local optimization. Each unlabeled client retains its local projection head across rounds. After aggregating client updates, the server updates the EMA teacher for the next round. The EMA checkpoint with the highest equally weighted mean validation accuracy across labeled source clients is retained for inference. DP-FedUDG follows the same procedure but replaces local optimization on labeled and unlabeled clients with DP-SGD, as defined in Eq.~\eqref{eq:dpsgd}. Each client's privacy accountant is updated accordingly. At inference, the selected EMA encoder and linear classifier predict $\hat y=\arg\max_j[\operatorname{softmax}(\bar W_{\mathrm{best}}\bar G_{\mathrm{best}}(x))]_j$.

\section{Experiments}
\label{sec:experiments}

This section evaluates FedUDG across four protocols and examines its component contributions, parameter sensitivity, and performance under differential privacy.

\subsection{Datasets and Evaluation Protocols}

FedUDG is evaluated on three visual-domain benchmarks. PACS \cite{li2017pacs} contains four domains, \texttt{Photo} (P), \texttt{Art Painting} (A), \texttt{Cartoon} (C), and \texttt{Sketch} (S), with seven shared classes. DomainNet \cite{peng2019domainnet} contains six domains and 345 classes; we use its seven-class subset aligned with PACS, denoted miniDomainNet, covering \texttt{Real} (R), \texttt{Painting} (P), \texttt{Clipart} (C), \texttt{Quickdraw} (Q), \texttt{Infograph} (I), and \texttt{Sketch} (S), where the DomainNet class \texttt{face} is mapped to the PACS class \texttt{person}. VLCS \cite{torralba2011vlcs} contains four domains, \texttt{VOC2007} (V), \texttt{LabelMe} (L), \texttt{Caltech101} (C), and \texttt{SUN09} (S), with five shared classes.
To assess the robustness of unseen-domain generalization across diverse federated settings, we consider four evaluation protocols. These protocols vary the source domains represented by labeled and unlabeled clients, the degree of within-domain class imbalance, and the coverage of labeled supervision:
\begin{enumerate}
    \item \textbf{PACS--DomainNet joint dataset:}
    Each PACS domain is held out in turn as the unseen target. The remaining PACS domains provide labeled clients, and all six miniDomainNet domains provide unlabeled clients. Each source domain forms one client or is randomly partitioned into two or four clients, yielding $(M_L,M_U)=(3,6)$, $(6,12)$, and $(12,24)$, respectively. An additional setting uses within-domain Dirichlet partitions with $\alpha=0.5$ and $(M_L,M_U)=(12,24)$. This protocol emphasizes feature shifts between labeled and unlabeled clients.

    \item \textbf{miniDomainNet with within-domain class imbalance:}
    Each of the six domains serves once as the unseen target. Each remaining source domain is partitioned into three disjoint clients through a class-wise Dirichlet allocation, with one client fully labeled and two fully unlabeled, giving $(M_L,M_U)=(5,10)$. We use $\alpha\in\{10,1,0.1\}$ to vary class-proportion heterogeneity. Both label roles are therefore represented within every source domain.

    \item \textbf{VLCS with within-domain class imbalance:}
    Each of the four domains serves once as the unseen target. The remaining three source domains follow the same partitioning and label-assignment procedure as miniDomainNet, yielding $(M_L,M_U)=(3,6)$ under $\alpha\in\{10,1,0.1\}$.

    \item \textbf{Single labeled-source domain:}
    Each source domain forms one client. X-Y denotes that X is the only labeled source client, Y is the unseen target client, and the remaining domains provide fully unlabeled source clients. For example, V-S uses \texttt{VOC2007} as the labeled source, \texttt{SUN09} as the target, and \texttt{LabelMe} and \texttt{Caltech101} as unlabeled sources. 
\end{enumerate}

In the four evaluation protocols, data of labeled clients is split into 80\% and 20\% for training and validation, respectively. 

\subsection{Implementation Details and Baselines}

\begin{table}[t]
\caption{Model and Training Configuration}
\label{tab:protocol}
\centering
\footnotesize
\setlength{\tabcolsep}{4.0pt}
\begin{tabular}{lc}
\toprule
\textbf{Parameter} & \textbf{Value} \\
\midrule
Backbone & ImageNet-pretrained ResNet-18 \\
Input resolution & $224\times224$ \\
Communication rounds & 80 \\
Local epochs & 5 \\
Batch size & 16 \\
Optimizer & SGD \\
Initial learning rate & $10^{-3}$ \\
Momentum & 0.9 \\
Weight decay & $5\times10^{-4}$ \\
LR schedule & $\gamma=0.1$ \\
Random seed & 1337, 1338, 1339 \\
\bottomrule
\end{tabular}
\end{table}
All experiments are implemented in PyTorch with Python~3.9 and run on an NVIDIA\textsuperscript{\textregistered} GeForce RTX 4090 GPU. Table~\ref{tab:protocol} summarizes the shared training settings, including the ImageNet-pretrained ResNet-18 backbone \cite{he2016resnet,deng2009imagenet}.
All experiments were implemented in PyTorch based on Python 3.9. The hardware platform consisted of an NVIDIA\textsuperscript{\textregistered} GeForce RTX 4090 GPU. FedUDG uses 10 supervised warm-up rounds, followed by one unlabeled local epoch per communication round. The base per-step EMA decay is $\rho_0=0.999$, converted to a round-level decay as described in Section~\ref{sec:method}. The confidence and JS thresholds for consensus-based sample admission are $\tau=0.8$ and $\kappa=0.1$. The anchor temperature $\beta_V$, anchor-loss weight $\lambda_V$, representation-distillation weight $\lambda_{\mathrm{Dis}}$, and maximum semantic-loss weight $\lambda_{\mathrm{sem}}^{\max}$ are set to 0.1, 1.0, 0.5, and 1.0. The maximum unlabeled mixing weight $\alpha_U^{\max}$ is 0.5 and increases linearly during the first 20 joint-training rounds. The encoder proximal coefficient is $\mu=10^{-3}$, and the hidden dimension of the local projection head $\Phi_u$ is 512.
The weak augmentation $\mathcal T_w$ consists of RandomResizedCrop with scale $[0.7,1.0]$, random horizontal flipping with probability 0.5, ColorJitter with magnitude 0.4, random grayscale conversion with probability 0.1, and ImageNet normalization. The strong augmentation $\mathcal T_s$ uses the same random crop and horizontal flip, followed by RandAugment \cite{cubuk2020randaugment} with two operations and magnitude 10, and then ImageNet normalization. FedUDG hyperparameters are set empirically and selected or checked using only labeled-source validation data. 

We compare FedUDG with eight baselines from two families. The FSSL baselines are FedIRM \cite{liu2021fedirm}, FedPDM \cite{liao2025fedpdm}, FedRVR \cite{yang2025fedrvr}, and RSCFed \cite{liang2022rscfed}. They share the client-level label topology of FedUDG, but they are designed for semi-supervised classification on the participating client distributions rather than OOD generalization to an unseen client. The FDG baselines are FedGA \cite{zhang2023feddgga}, FedAlign \cite{gupta2025fedalign}, FedDIM \cite{liao2025decision}, and FedIIR \cite{guo2023invariant}. They target generalization to unseen domains but assume that every source client is fully labeled, and are therefore trained only on the labeled source clients. The two families thus provide complementary references: FSSL methods use unlabeled clients but lack an explicit OOD objective, whereas FDG methods pursue OOD generalization but are restricted to labeled data. FedUDG addresses both aspects in a single framework. To control for model capacity and training budget, all methods use the same backbone and the shared settings in Table~\ref{tab:protocol}, while method-specific objectives, augmentations, and hyperparameters follow the original papers or public implementations. We report the classification accuracy on the unseen target client, averaged over three independent runs.

\subsection{Main Results}
\begin{table*}[t]
\caption{Classification Accuracy (\%) on Held-Out PACS Targets in the PACS--DomainNet Joint Dataset. $M=(M_L,M_U)$ denotes the labeled and unlabeled client counts. Best and second-best results are bold and underlined, respectively.}
\label{tab:joint}
\centering
\footnotesize
\setlength{\tabcolsep}{0.5pt}
\setlength{\aboverulesep}{0pt}
\setlength{\belowrulesep}{0pt}
\setlength{\extrarowheight}{1.5pt}
\begin{tabular*}{\textwidth}{@{\extracolsep{\fill}}lccccg@{\hspace{4pt}}ccccg@{\hspace{4pt}}ccccg@{\hspace{4pt}}ccccg}
\toprule
& \multicolumn{5}{c}{\textbf{\boldmath$M=(3,6)$}} & \multicolumn{5}{c}{\textbf{\boldmath$M=(6,12)$}} & \multicolumn{5}{c}{\textbf{\boldmath$M=(12,24)$}} & \multicolumn{5}{c}{\textbf{\boldmath$M=(12,24),\ \alpha=0.5$}} \\
\cmidrule(lr){2-6}\cmidrule(lr){7-11}\cmidrule(lr){12-16}\cmidrule(lr){17-21}
\textbf{Method} & \textbf{P} & \textbf{A} & \textbf{C} & \textbf{S} & \multicolumn{1}{c@{\hspace{4pt}}}{\textbf{Mean}} & \textbf{P} & \textbf{A} & \textbf{C} & \textbf{S} & \multicolumn{1}{c@{\hspace{4pt}}}{\textbf{Mean}} & \textbf{P} & \textbf{A} & \textbf{C} & \textbf{S} & \multicolumn{1}{c@{\hspace{4pt}}}{\textbf{Mean}} & \textbf{P} & \textbf{A} & \textbf{C} & \textbf{S} & \multicolumn{1}{c}{\textbf{Mean}} \\
\midrule
FedIRM \cite{liu2021fedirm} & 93.41 & 73.14 & 72.78 & 74.21 & 78.39 & 92.57 & 73.71 & 72.01 & 73.56 & 77.96 & 90.16 & 72.68 & 71.17 & 72.59 & 76.65 & 87.51 & 71.93 & 68.12 & 62.24 & 72.45 \\
FedPDM \cite{liao2025fedpdm} & 91.02 & 76.46 & 77.60 & 79.15 & 81.06 & 87.66 & 73.49 & 75.73 & 64.95 & 75.46 & 86.11 & 64.75 & 72.10 & 45.74 & 67.18 & 81.98 & 58.15 & 68.26 & 39.35 & 61.94 \\
FedRVR \cite{yang2025fedrvr} & \textbf{95.69} & 81.15 & 74.57 & 65.69 & 79.28 & \underline{93.89} & 79.98 & 74.70 & 59.00 & 76.89 & 93.15 & 76.98 & 70.59 & 56.21 & 74.23 & 92.34 & 74.35 & 65.23 & 45.29 & 69.30 \\
RSCFed \cite{liang2022rscfed} & 91.44 & 75.39 & \underline{78.11} & 72.00 & 79.24 & 92.93 & 77.25 & 75.64 & 70.88 & 79.18 & \underline{93.83} & 75.44 & \underline{75.17} & 61.64 & 76.52 & \underline{93.89} & 69.68 & 70.56 & 60.75 & 73.72 \\
\midrule
FedGA \cite{zhang2023feddgga} & 93.86 & 81.27 & 76.84 & 82.83 & 83.70 & 91.41 & 79.82 & 75.18 & 83.23 & 82.41 & 91.08 & 77.27 & 73.97 & 80.45 & 80.69 & 90.95 & 74.23 & 71.62 & 79.90 & 79.18 \\
FedAlign \cite{gupta2025fedalign} & 94.35 & 79.13 & 76.74 & 78.17 & 82.10 & 92.24 & 78.50 & \underline{76.99} & 75.21 & 80.74 & 91.59 & 76.42 & 74.17 & 75.59 & 79.44 & 88.54 & 73.39 & \underline{74.14} & 74.59 & 77.67 \\
FedDIM \cite{liao2025decision} & 94.21 & 81.96 & 74.12 & \textbf{86.41} & \underline{84.18} & 93.36 & 81.33 & 74.24 & \textbf{86.15} & \underline{83.77} & 92.09 & \underline{80.21} & 73.10 & \textbf{84.87} & \underline{82.57} & 90.35 & 79.20 & 72.34 & \underline{81.27} & \underline{80.79} \\
FedIIR \cite{guo2023invariant} & 93.36 & \underline{84.02} & 72.24 & 83.59 & 83.30 & 90.77 & \underline{82.14} & 71.79 & 82.06 & 81.69 & 90.16 & 79.76 & 72.08 & 81.20 & 80.80 & 86.19 & \underline{80.02} & 70.08 & 80.06 & 79.09 \\
\midrule
\textbf{FedUDG} & \underline{95.63} & \textbf{85.69} & \textbf{78.75} & \underline{86.21} & \textbf{86.57} & \textbf{94.97} & \textbf{86.04} & \textbf{78.41} & \underline{85.37} & \textbf{86.20} & \textbf{95.75} & \textbf{85.40} & \textbf{76.75} & \underline{84.34} & \textbf{85.56} & \textbf{94.55} & \textbf{83.89} & \textbf{74.70} & \textbf{84.04} & \textbf{84.30} \\
\bottomrule
\end{tabular*}
\end{table*}

\begin{table*}[t]
\caption{Classification Accuracy (\%) on miniDomainNet under Within-Domain Dirichlet Partitions with $(M_L,M_U)=(5,10)$. Best and second-best results are bold and underlined, respectively.}
\label{tab:minidomainnet_dirichlet}
\centering
\footnotesize
\setlength{\tabcolsep}{0.5pt}
\setlength{\aboverulesep}{0pt}
\setlength{\belowrulesep}{0pt}
\setlength{\extrarowheight}{1.5pt}
\begin{tabular*}{\textwidth}{@{\extracolsep{\fill}}l*{6}{w{c}{18pt}}>{\columncolor{gray!12}}w{c}{18pt}*{6}{w{c}{18pt}}>{\columncolor{gray!12}}w{c}{18pt}*{6}{w{c}{18pt}}>{\columncolor{gray!12}}w{c}{18pt}}
\toprule
& \multicolumn{7}{c}{\textbf{\boldmath$\alpha=10$}} & \multicolumn{7}{c}{\textbf{\boldmath$\alpha=1$}} & \multicolumn{7}{c}{\textbf{\boldmath$\alpha=0.1$}} \\
\cmidrule(lr){2-8}\cmidrule(lr){9-15}\cmidrule(lr){16-22}
\textbf{Method} & \textbf{S} & \textbf{R} & \textbf{Q} & \textbf{C} & \textbf{P} & \textbf{I} & \textbf{Mean} & \textbf{S} & \textbf{R} & \textbf{Q} & \textbf{C} & \textbf{P} & \textbf{I} & \textbf{Mean} & \textbf{S} & \textbf{R} & \textbf{Q} & \textbf{C} & \textbf{P} & \textbf{I} & \textbf{Mean} \\
\midrule
FedIRM \cite{liu2021fedirm} & 76.54 & 81.97 & 33.25 & 86.30 & 85.18 & 43.35 & 67.77 & 76.01 & 81.52 & 32.75 & 85.67 & 85.03 & 41.36 & 67.06 & 70.19 & 72.21 & 30.25 & 68.15 & 80.42 & 34.20 & 59.24 \\
FedPDM \cite{liao2025fedpdm} & 78.72 & 83.78 & 33.86 & 88.15 & 83.14 & 45.39 & 68.84 & 77.65 & 84.51 & 33.26 & 87.58 & 84.09 & 44.12 & 68.54 & 69.74 & 74.38 & 31.98 & 70.36 & 82.29 & 38.53 & 61.21 \\
FedRVR \cite{yang2025fedrvr} & 80.97 & 93.13 & 37.09 & 91.72 & 86.95 & 46.45 & 72.72 & 79.57 & 92.82 & 35.99 & 87.23 & 85.82 & 44.15 & 70.93 & 62.36 & 88.95 & 38.21 & 69.20 & 79.41 & 47.09 & 64.20 \\
RSCFed \cite{liang2022rscfed}  & 81.19 & 89.22 & 34.57 & 90.61 & 85.17 & 45.11 & 70.98 & 80.90 & 87.50 & 36.51 & 89.10 & 87.00 & 44.55 & 70.93 & 61.50 & 55.10 & 39.00 & 70.90 & 76.80 & \underline{48.20} & 58.58 \\
\midrule
FedGA \cite{zhang2023feddgga} & 84.01 & \underline{94.08} & 56.29 & \underline{92.61} & 88.62 & 43.98 & 76.60 & 82.70 & \underline{93.95} & 54.71 & 89.33 & 87.91 & 43.25 & 75.31 & 64.26 & 87.19 & \underline{49.04} & \underline{73.58} & 85.92 & 42.59 & 67.10 \\
FedAlign \cite{gupta2025fedalign} & \underline{86.44} & 93.17 & 52.91 & \underline{92.61} & 91.05 & \textbf{50.88} & 77.84 & 84.28 & 93.17 & 50.26 & \textbf{91.43} & \underline{90.21} & \underline{50.02} & 76.56 & 76.53 & \underline{89.12} & 45.78 & 72.83 & \textbf{88.50} & 44.39 & 69.53 \\
FedDIM \cite{liao2025decision} & 85.63 & 93.86 & 57.60 & 91.23 & 90.67 & \underline{50.58} & 78.26 & \underline{85.11} & 93.91 & 55.78 & 88.32 & 88.53 & 48.05 & 76.62 & 79.45 & \textbf{90.18} & 47.06 & 68.15 & \underline{88.10} & 47.43 & \underline{70.06} \\
FedIIR \cite{guo2023invariant} & \textbf{86.99} & 93.15 & \textbf{60.46} & 90.96 & \underline{91.19} & 50.25 & \underline{78.83} & 84.65 & 93.06 & \textbf{59.52} & 88.27 & 89.43 & \textbf{50.56} & \underline{77.58} & \underline{80.21} & 85.30 & 45.97 & 66.99 & 83.59 & \textbf{48.22} & 68.38 \\
\midrule
\textbf{FedUDG} & 86.14 & \textbf{94.86} & \underline{58.26} & \textbf{93.25} & \textbf{91.71} & 50.25 & \textbf{79.08} & \textbf{86.09} & \textbf{94.68} & \underline{57.99} & \underline{90.67} & \textbf{90.83} & 47.25 & \textbf{77.92} & \textbf{81.22} & 87.79 & \textbf{53.74} & \textbf{74.65} & 86.05 & 42.22 & \textbf{70.95} \\
\bottomrule
\end{tabular*}
\end{table*}

\begin{figure*}[!t]
    \centering
    \includegraphics[width=1\textwidth]{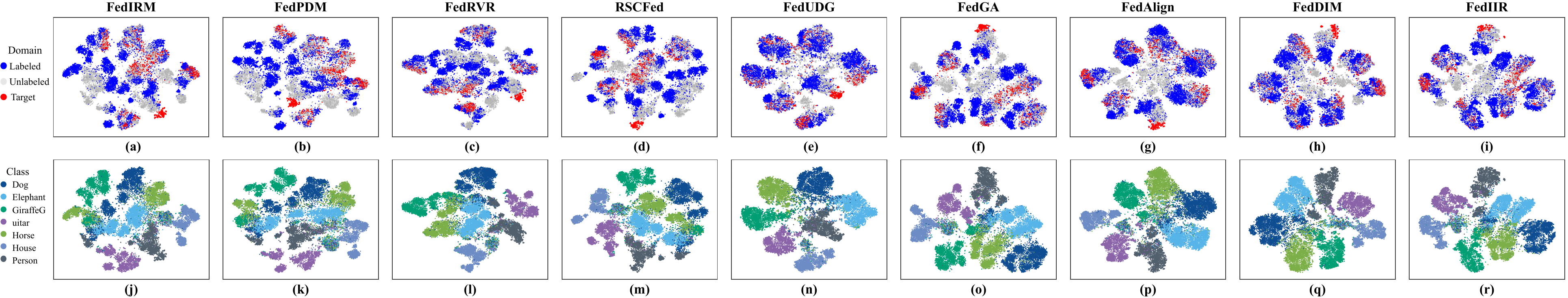}
    \caption{t-SNE visualization of encoder features for the Photo target in the PACS--DomainNet joint dataset ($M=(6,12)$).}
    \label{fig:client18_tsne}
\end{figure*}

Table~\ref{tab:joint} reports the results on the PACS--DomainNet joint dataset. FedUDG achieves the best mean accuracy in all four settings and the best result in 16 of the 20 reported columns. It outperforms the strongest baseline, FedDIM, by 2.39, 2.43, 2.99, and 3.51 percentage points, and the strongest FSSL baseline by 5.51, 7.02, 8.91, and 10.58 points. As the federation grows larger and more heterogeneous, FedUDG stays more stable than the baselines, and its advantage widens. From $M=(3,6)$ to $M=(12,24),\ \alpha=0.5$, FedUDG drops by only 2.27 points, whereas the FDG and FSSL baselines drop by 3.39--4.52 and 5.52--19.12 points, respectively. Notably, every FDG baseline outperforms every FSSL baseline in mean accuracy in all settings, even though the FDG baselines use no unlabeled data. The gap between the best methods of the two groups also widens from 3.12 to 7.07 points as $M_U$ increases. Access to additional unlabeled clients alone therefore does not ensure better OOD generalization. FedUDG outperforms all evaluated baselines under this protocol, supporting the effectiveness of its controlled semantic transfer across labeled and unlabeled domains.
Fig.~\ref{fig:client18_tsne} offers qualitative insights into the FDG--FSSL performance gap. FedUDG and the FDG baselines tend to align classes across domains and form tighter within-class clusters of source and target samples. In contrast, FSSL appears to struggle with cross-domain alignment, often fragmenting instances of the same class into separate subclusters. This pattern suggests that domain alignment is an important factor for OOD generalization.

\begin{table*}[t]
\caption{Classification Accuracy (\%) on VLCS under Within-Domain Dirichlet Partitions with $(M_L,M_U)=(3,6)$. Best and second-best results are bold and underlined, respectively.}
\label{tab:vlcs_dirichlet}
\centering
\footnotesize
\setlength{\tabcolsep}{1.0pt}
\setlength{\aboverulesep}{0.5pt}
\setlength{\belowrulesep}{0pt}
\setlength{\extrarowheight}{0pt}
\begin{tabular*}{\textwidth}{@{\extracolsep{\fill}}lccccg@{\hspace{7pt}}ccccg@{\hspace{7pt}}ccccg}
\toprule
& \multicolumn{5}{c}{\textbf{\boldmath$\alpha=10$}} & \multicolumn{5}{c}{\textbf{\boldmath$\alpha=1$}} & \multicolumn{5}{c}{\textbf{\boldmath$\alpha=0.1$}} \\
\cmidrule(lr){2-6}\cmidrule(lr){7-11}\cmidrule(lr){12-16}
\textbf{Method} & \textbf{V} & \textbf{L} & \textbf{C} & \textbf{S} & \multicolumn{1}{c@{\hspace{7pt}}}{\textbf{Mean}} & \textbf{V} & \textbf{L} & \textbf{C} & \textbf{S} & \multicolumn{1}{c@{\hspace{7pt}}}{\textbf{Mean}} & \textbf{V} & \textbf{L} & \textbf{C} & \textbf{S} & \textbf{Mean} \\
\midrule
FedIRM \cite{liu2021fedirm} & 62.10 & 51.57 & 91.81 & 67.92 & 68.35  & 47.83 & 53.52 & 90.16 & 66.58 & 64.52 & 36.49 & 50.17 & 87.32 & 51.21 & 56.30 \\
FedRVR \cite{yang2025fedrvr} & 62.34 & 51.26 & \underline{97.17} & 58.86 & 67.41 & 61.63 & 50.98 & \underline{95.71} & 55.85 & 66.04 & 31.29 & 48.42 & 94.54 & 44.67 & 54.73  \\
FedPDM \cite{liao2025fedpdm} & 65.72 & 53.09 & 93.44 & 68.57 & 70.21 & 63.15 & 49.11 & 93.24 & 66.19 & 67.92 &  35.32 & 48.27 & 83.30 & 49.16 & 54.01 \\
RSCFed \cite{liang2022rscfed} & 66.32 & \underline{59.45} & 95.95 & 68.72 & 72.61 & 58.90 & \textbf{60.34} & \textbf{96.15} & 68.06 & 70.86 & 35.22 & 55.47 & 92.09 & \textbf{62.67} & \underline{61.36} \\
\midrule
FedGA \cite{zhang2023feddgga} & 68.81 & 55.89 & 94.35 & 70.70 & 72.44 & 65.82 & 54.98 & 93.45 & 68.67 & 70.73 & 35.62 & 49.63 & 89.14 & 52.48 & 56.72 \\
FedAlign \cite{gupta2025fedalign} & 68.94 & 54.98 & 96.47 & 71.40 & 72.95 & 66.94 & 56.03 & 94.10 & \underline{69.60} & 71.67 & 36.40 & 50.29 & 90.68 & 54.34 & 57.93 \\
FedDIM \cite{liao2025decision} & \textbf{71.18} & 57.24 & \textbf{97.28} & 70.83 & 74.13 & \underline{68.01} & 58.88 & 95.44 & 68.09 & 72.61 & 36.35 & 55.94 & \underline{95.06} & 53.42 & 60.19 \\
FedIIR \cite{guo2023invariant} & \underline{70.79} & 58.49 & 95.86 & \textbf{72.01} & \underline{74.29} & \textbf{68.17} & 58.32 & 95.59 & \textbf{70.94} & \textbf{73.26} & \underline{37.20} & \underline{57.16} & 91.43 & \underline{58.28} & 61.02 \\
\midrule
\textbf{FedUDG} & 68.06 & \textbf{60.81} & \underline{97.17} & \underline{71.57} & \textbf{74.40} & 67.88 & \underline{60.11} & 95.15 & 69.09 & \underline{73.06} & \textbf{39.86} & \textbf{59.44} & \textbf{95.22} & 56.86 & \textbf{62.85} \\
\bottomrule
\end{tabular*}
\end{table*}

\begin{table*}[t]
\caption{Classification Accuracy (\%) with One Labeled Source Client on PACS, miniDomainNet, and VLCS. Best and second-best results are bold and underlined, respectively.}
\label{tab:three_datasets}
\centering
\footnotesize
\setlength{\tabcolsep}{1.0pt}
\setlength{\aboverulesep}{0pt}
\setlength{\belowrulesep}{0pt}
\setlength{\extrarowheight}{1.5pt}
\begin{tabular*}{\textwidth}{@{\extracolsep{\fill}}lccccg@{\hspace{7pt}}ccccccg@{\hspace{7pt}}ccccg}
\toprule
& \multicolumn{5}{c}{\textbf{PACS}} & \multicolumn{7}{c}{\textbf{miniDomainNet}} & \multicolumn{5}{c}{\textbf{VLCS}} \\
\cmidrule(lr){2-6}\cmidrule(lr){7-13}\cmidrule(lr){14-18}
\textbf{Method} & \textbf{P-S} & \textbf{A-P} & \textbf{C-A} & \textbf{S-C} & \multicolumn{1}{c@{\hspace{7pt}}}{\textbf{Mean}} & \textbf{R-S} & \textbf{P-R} & \textbf{C-Q} & \textbf{S-C} & \textbf{I-P} & \textbf{Q-I} & \multicolumn{1}{c@{\hspace{7pt}}}{\textbf{Mean}} & \textbf{V-S} & \textbf{L-V} & \textbf{C-L} & \textbf{S-C} & \multicolumn{1}{c}{\textbf{Mean}} \\
\midrule
FedIRM \cite{liu2021fedirm} & 28.81 & 86.96 & 63.81 & 49.43 & 57.25 & 78.78 & 89.82 & \underline{55.05} & 71.61 & 64.52 & 17.60 & 62.90 & 68.12 & 56.36 & 46.39 & 59.48 & 57.59 \\
FedPDM \cite{liao2025fedpdm} & 27.38 & 94.67 & \underline{74.95} & 44.11 & 60.28 & 79.93 & 93.97 & 33.65 & \textbf{87.55} & \textbf{89.09} & \underline{23.39} & \underline{67.93} & \textbf{74.84} & 60.81 & 37.17 & 60.54 & 58.34 \\
FedRVR \cite{yang2025fedrvr} & \textbf{41.74} & 92.93 & 66.30 & 47.52 & \underline{62.12} & \underline{81.64} & 91.86 & 47.57 & 83.12 & 43.76 & 15.13 & 60.51 & 71.83 & 52.26 & 23.37 & \textbf{76.59} & 56.01 \\
RSCFed \cite{liang2022rscfed} & 29.19 & \underline{95.80} & 64.74 & \underline{57.59} & 61.83 & 79.81 & \textbf{95.47} & 42.37 & 80.24 & 71.95 & 20.41 & 65.04 & \underline{74.27} & \underline{61.63} & \textbf{57.67} & 61.55 & \underline{63.78} \\
\textbf{FedUDG} & \underline{35.99} & \textbf{96.04} & \textbf{76.61} & \textbf{67.02} & \textbf{68.92} & \textbf{82.13} & \underline{94.19} & \textbf{62.38} & \underline{85.09} & \underline{85.19} & \textbf{23.85} & \textbf{72.14} & 74.01 & \textbf{63.44} & \underline{52.56} & \underline{70.64} & \textbf{65.16} \\
\bottomrule
\end{tabular*}
\end{table*}

Tables~\ref{tab:minidomainnet_dirichlet} and~\ref{tab:vlcs_dirichlet} report the results under within-domain Dirichlet partitions. On miniDomainNet, FedUDG achieves mean accuracies of 79.08\%, 77.92\%, and 70.95\% for $\alpha=10$, $1$, and $0.1$, respectively. These results exceed the strongest FDG baseline in each setting by 0.25, 0.34, and 0.89 percentage points. The gains over the strongest FSSL baseline range from 6.36 to 6.99 points. On VLCS, FedUDG reaches 74.40\%, 73.06\%, and 62.85\% under the same three $\alpha$ values. It leads the strongest baseline by 0.11 points at $\alpha=10$ and 1.49 points at $\alpha=0.1$, but trails FedIIR by 0.20 points at $\alpha=1$. FedUDG thus achieves the highest mean accuracy in five of the six settings across the two datasets. Note that to focus on exploring how unlabeled clients enhance OOD generalization, FedUDG does not employ standard FDG strategies for labeled clients, suggesting synergistic potential with existing FDG methods. However, decreasing $\alpha$ from 10 to 0.1 reduces its mean accuracy by 8.13 points on miniDomainNet and 11.55 points on VLCS. Severe class imbalance therefore remains challenging despite its competitive relative performance.

Table~\ref{tab:three_datasets} compares FedUDG with the FSSL baselines under a single labeled-source setting. FedUDG achieves the highest mean accuracy on PACS, miniDomainNet, and VLCS (68.92\%, 72.14\%, and 65.16\%), exceeding the strongest baseline by 6.80, 4.21, and 1.38 percentage points. The strongest baseline also differs across datasets (FedRVR, FedPDM, and RSCFed). The largest gains arise under large appearance shifts, such as S-C on PACS (+9.43) and C-Q on miniDomainNet (+7.33). Across the 14 tasks, FedUDG ranks first on 7, second on 6, and third on V-S, where it is 0.83 points below the best result. The baselines are markedly less stable. For example, FedRVR is best on VLCS S-C but reaches only 23.37\% on C-L, and FedPDM is best on miniDomainNet I-P but reaches only 33.65\% on C-Q. FedUDG still trails the best baseline by 3.90--5.95 points on PACS P-S, miniDomainNet I-P, and VLCS C-L and S-C, indicating that a single labeled source can limit its advantage on certain source--target pairs.

\subsection{Ablation Study}

\begin{figure}[!t]
    \centering
    \includegraphics[width=0.98\linewidth]{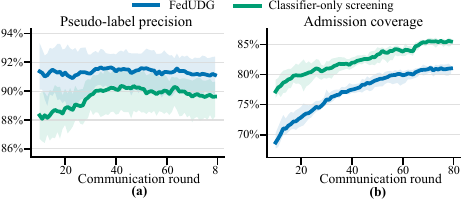}
    \caption{Pseudo-label precision among admitted samples (a) and admission coverage (b) on the PACS–miniDomainNet joint dataset. Lines show the mean across four target tasks; shaded bands show the minimum–maximum range.}
    \label{fig:Pseudo-label}
\end{figure}

\begin{table}[t]
\caption{Ablation Results (\%) on Held-Out PACS Targets in the PACS--DomainNet Joint Dataset with $M=(3,6)$.}
\label{tab:ablation}
\centering
\footnotesize
\setlength{\tabcolsep}{3.0pt}
\begin{tabular*}{\columnwidth}{@{\extracolsep{\fill}}lccccc}
\toprule
\textbf{Variant} & \textbf{P} & \textbf{A} & \textbf{C} & \textbf{S} & \textbf{Mean} \\
\midrule
FedAvg & 92.56 & 77.08 & 77.87 & 81.11 & 82.15 \\
FedUDG w/o unlabeled & 92.29 & 80.84 & 76.56 & 82.81 & 83.13 \\
\midrule
FedUDG w/o $\mathcal L_{\mathrm{sem}}$ & 94.01 & 83.01 & \textbf{78.97} & 84.58 & 85.14 \\
FedUDG w/o $\mathcal L_{\mathrm{Dis}}$ & 95.51 & \underline{85.24} & 77.99 & 84.86 & \underline{85.90} \\
FedUDG w/o JS & \textbf{96.07} & 84.68 & 77.01 & \underline{85.28} & 85.76 \\
FedUDG w/o anchor \& JS & 95.43 & 82.56 & 77.66 & 84.87 & 85.13 \\
\midrule
\textbf{FedUDG} & \underline{95.63} & \textbf{85.69} & \underline{78.75} & \textbf{86.21} & \textbf{86.57} \\
\bottomrule
\end{tabular*}
\end{table}
FedAvg trains only on labeled source clients with a linear classifier, whereas FedUDG w/o unlabeled retains FedUDG's supervised training with semantic anchors but excludes unlabeled-client updates. As shown in Table~\ref{tab:ablation}, full FedUDG reaches a mean accuracy of 86.57\%, exceeding these two variants by 4.42 and 3.44 percentage points, respectively. The latter gap isolates the contribution of unlabeled sources beyond the labeled-path design. Removing semantic self-training or label-free representation distillation lowers the mean accuracy to 85.14\% and 85.90\%, respectively; the larger drop from removing $\mathcal L_{\mathrm{sem}}$ suggests that class-level supervision on admitted samples is the main source of the unlabeled-client gain, while distillation provides a complementary improvement. For consensus-based sample admission, FedUDG w/o JS retains the classifier--anchor agreement check, whereas FedUDG w/o anchor \& JS further uses a single weak view with classifier confidence only. Their mean accuracies decrease to 85.76\% and 85.13\%, respectively.
To examine screening beyond target accuracy, we track pseudo-label precision and admission coverage during training. Let $\mathcal A_t=\{x:m_t(x)=1\}$ denote the samples admitted at round $t$. Precision $\operatorname{Prec}_t$ is the fraction of $x\in\mathcal A_t$ whose soft target satisfies $\arg\max_j q_j(x)=y_x$, where $y_x$ is the ground-truth label used only for this analysis, and coverage is $\operatorname{Cov}_t=|\mathcal A_t|/\sum_{u\in\mathcal C_U}|\mathcal S_u^U|$. Over the last five joint-training rounds, full FedUDG and FedUDG w/o anchor \& JS reach mean precision of 91.14\% and 89.65\% and mean coverage of 80.96\% and 85.47\%, respectively (Fig.~\ref{fig:Pseudo-label} ). The full rule thus trades 4.51 points of coverage for 1.49 points of precision, yielding cleaner semantic supervision.

\subsection{Parameter Sensitivity}

We study three parameters introduced by FedUDG: the confidence threshold $\tau$, the JS threshold $\kappa$, and the maximum unlabeled mixing weight $\alpha_U^{\max}$. Using the PACS--DomainNet joint dataset with $M=(3,6)$, we vary one parameter at a time over $\tau\in\{0.7,0.8,0.9\}$, $\kappa\in\{0.05,0.1,0.2\}$, and $\alpha_U^{\max}\in\{0.3,0.5,0.7\}$, keeping the others at their defaults $(0.8,0.1,0.5)$. This analysis is performed post hoc, and target accuracy is deliberately excluded from hyperparameter selection.
As shown in Table~\ref{tab:sensitivity}, mean accuracy ranges from 85.66\% to 86.57\% across all settings, a maximum variation of 0.91 percentage points. Even the lowest setting exceeds the strongest baseline at $M=(3,6)$ by 4.60 points, suggesting that the advantage of FedUDG does not depend on precise tuning. For each parameter, the highest accuracy is achieved at its default value. For $\tau$, the stricter threshold of 0.9 lowers the mean by 0.76 points, more than the 0.39-point drop at 0.7, consistent with the observation in ablation study that excessive filtering reduces coverage without a commensurate precision gain. For $\kappa$, values of 0.05 and 0.2 reduce the mean by 0.46 and 0.83 points. For $\alpha_U^{\max}$, a value of 0.3 gives the lowest mean of 85.66\%, indicating that unlabeled clients require sufficient aggregation weight; increasing it to 0.7 improves Art Painting to 86.96\% but lowers Cartoon to 76.84\%, yielding a slightly lower mean.
\begin{table}[t]
\caption{Parameter Sensitivity (\%) on Held-Out PACS Targets in the PACS--DomainNet Joint Dataset with $M=(3,6)$.}
\label{tab:sensitivity}
\centering
\footnotesize
\setlength{\tabcolsep}{3.0pt}
\begin{tabular*}{\columnwidth}{@{\extracolsep{\fill}}lcccccc}
\toprule
\textbf{Parameter} & \textbf{Value} & \textbf{P} & \textbf{A} & \textbf{C} & \textbf{S} & \textbf{Mean} \\
\midrule
\multirow{3}{*}{$\tau$} & 0.7 & 95.51 & 85.01 & \textbf{79.18} & 85.03 & 86.18 \\
 & 0.8 & \textbf{95.63} & \textbf{85.69} & 78.75 & \textbf{86.21} & \textbf{86.57} \\
 & 0.9 & 94.97 & 84.13 & 78.71 & 85.42 & 85.81 \\
\midrule
\multirow{3}{*}{$\kappa$} & 0.05 & 94.85 & \textbf{85.74} & 78.71 & 85.14 & 86.11 \\
 & 0.1 & \textbf{95.63} & 85.69 & \textbf{78.75} & \textbf{86.21} & \textbf{86.57} \\
 & 0.2 & 94.92 & 84.90 & 78.09 & 85.03 & 85.74 \\
\midrule
\multirow{3}{*}{$\alpha_U^{\max}$} & 0.3 & 94.82 & 84.84 & 78.03 & 84.93 & 85.66 \\
 & 0.5 & \textbf{95.63} & 85.69 & \textbf{78.75} & \textbf{86.21} & \textbf{86.57} \\
 & 0.7 & 95.55 & \textbf{86.96} & 76.84 & 85.53 & 86.22 \\
\bottomrule
\end{tabular*}
\end{table}

\subsection{Privacy Evaluation}

\begin{table}[t]
\caption{Accuracy (\%) on Held-Out PACS Targets under Sample-Level DP in the PACS--DomainNet Joint Dataset ($\delta=10^{-5}$, $C=1$).}
\label{tab:privacy}
\centering
\footnotesize
\setlength{\tabcolsep}{2pt}
\begin{tabular*}{\columnwidth}{@{\extracolsep{\fill}}lccccccc}
\toprule
\textbf{Method} & \textbf{\boldmath$\sigma$} & \textbf{\boldmath$\varepsilon_{\max}$} & \textbf{P} & \textbf{A} & \textbf{C} & \textbf{S} & \textbf{Mean} \\
\midrule
FedUDG & -- & $\infty$ & 95.63 & 85.69 & 78.75 & 86.21 & 86.57 \\
\midrule
\multirow{3}{*}{DP-FedUDG} & 1.0 & 16.94 & 73.59 & 53.11 & 42.92 & 57.25 & 56.72 \\
 & 1.5 & 8.46 & 53.95 & 39.51 & 34.02 & 41.72 & 42.30 \\
 & 2.0 & 5.66 & 29.46 & 20.34 & 22.99 & 25.35 & 24.54 \\
\bottomrule
\end{tabular*}
\end{table}

We evaluate DP-FedUDG on the PACS--DomainNet joint dataset with $M=(3,6)$, fixing $\delta=10^{-5}$ and clipping norm $C=1$ and varying the noise multiplier $\sigma\in\{1.0,1.5,2.0\}$. For each $\sigma$, the reported $\varepsilon_{\max}$ is the maximum privacy budget over all clients and the four target settings. The reported budget covers training records only. Checkpoint selection uses held-out labeled-source validation records, which are not included in this accounting.
As shown in Table~\ref{tab:privacy}, DP-FedUDG retains some unseen-domain utility at lower noise levels, but remains substantially below non-private FedUDG. At $\sigma=1.0$ ($\varepsilon_{\max}=16.94$), the mean accuracy is 56.72\%, a reduction of 29.85 percentage points. Increasing $\sigma$ to 1.5 and 2.0 reduces $\varepsilon_{\max}$ to 8.46 and 5.66, while mean accuracy falls to 42.30\% and 24.54\%, respectively. At the highest evaluated noise level, target accuracies range from 20.34\% to 29.46\%, indicating severely limited unseen-domain utility.
The degradation may reflect the combined effects of per-sample clipping, Gaussian noise, and frozen BatchNorm statistics. Perturbed local updates may reduce the reliability of the EMA teacher and the supervised semantic reference used for unlabeled-client learning. Under the evaluated configuration, DP-FedUDG does not yet achieve a satisfactory balance between sample-level privacy and OOD generalization, especially at higher noise levels. This remains a limitation of the current extension despite FedUDG's strong non-private performance.

\section{Conclusion}
\label{sec:conclusion}

This paper studied federated domain generalization with unlabeled clients. We proposed FedUDG to address heterogeneous label availability and improve OOD generalization. At the federation level, FedUDG uses a parameter-functional differentiation strategy. It maintains an additional semantic anchor supervised solely by ground-truth labels to protect class semantics, while the remaining parameters absorb knowledge from unlabeled clients through bounded aggregation. Within each unlabeled client, the student-teacher framework connects the protected semantic reference to unlabeled learning through two complementary objectives. Consensus between the classifier and label-supervised anchor provides soft targets for semantic self-training on credible samples. Label-free representation distillation exploits all unlabeled samples. Together, these objectives combine selective class-semantic guidance with feature stability under appearance perturbations to support generalization to unseen domains. 
On the PACS--DomainNet joint dataset, FedUDG achieved the highest mean target accuracy in all four settings, exceeding the strongest FSSL baseline by 5.51--10.58 percentage points and the strongest FDG baseline by 2.39--3.51 points. In the single-labeled-source setting on PACS, miniDomainNet, and VLCS, it exceeded the strongest baseline by 6.80, 4.21, and 1.38 points, respectively. Within-domain Dirichlet partitions further yielded the highest mean accuracy in five of six settings across two datasets. Ablations confirmed that unlabeled sources contribute gains beyond the labeled-path design.

We further evaluated DP-FedUDG, a sample-level differentially private extension. It retained some generalization ability at lower noise levels, but mean accuracy fell from 56.72\% to 24.54\% as $\sigma$ increased from 1.0 to 2.0. Under these settings, the extension did not preserve competitive OOD performance, particularly at stronger noise levels. DP-FedUDG therefore does not yet offer a satisfactory privacy--generalization trade-off. Future work will focus on preserving generalizable representations under formal, quantitatively accounted privacy constraints.

\bibliographystyle{IEEEtran}
\bibliography{main}

\makeatletter
\def\@IEEEBIOskipN{1\baselineskip}
\makeatother

\end{document}